\documentclass[runningheads]{llncs}

\usepackage{eccvabbrv}

\usepackage{graphicx}
\usepackage{booktabs}
\usepackage{arydshln}
\usepackage{textcomp}
\usepackage[accsupp]{axessibility}  % Improves PDF readability for those with disabilities.

\usepackage[pagebackref,breaklinks,colorlinks,citecolor=blue]{hyperref}
\usepackage{orcidlink}

\begin{document}

\title{MedXChat: A Unified Multimodal Large Language Model Framework towards CXRs Understanding and Generation} 

\titlerunning{MedXChat}

\author{Ling Yang\inst{1} \and
Zhanyu Wang\inst{1*} \and
Zhenghao Chen\inst{1} \and
Xinyu Liang\inst{2} \and
Luping Zhou\inst{1}\thanks{Corresponding author.}
}

\authorrunning{Ling Yang, Zhanyu Wang et al.}

\institute{The University of Sydney, Australia \\
\email{\{ling.yang2, zhanyu.wang, zhenghao.chen, luping.zhou\}@sydney.edu.au} \and
Guangzhou University of Chinese Medicine, China\\
\email{xinyu.liang31@gmail.com}}

\maketitle

\begin{abstract}
  % Despite the success of Large Language Models (LLMs) in general image tasks, a gap persists in the medical field for a multimodal large model adept at handling the nuanced diversity of medical images. Addressing this, we propose MedXChat, a unified multimodal large model designed for seamless interactions between medical assistants and users. MedXChat encompasses three key functionalities: CXR(Chest X-ray)-to-Report generation, CXR-based visual question-answering (VQA), and Text-to-CXR synthesis. Our contributions are as follows. Firstly, our model showcases exceptional cross-task adaptability, displaying adeptness across all three defined tasks and outperforming the benchmark models on the MIMIC dataset in medical multimodal applications. Secondly, we introduce an innovative Text-to-CXR synthesis approach that utilizes instruction-following capabilities within the Stable Diffusion (SD) architecture. This technique integrates smoothly with the existing model framework, requiring no extra parameters, thereby maintaining the SD's generative strength while also bestowing upon it the capacity to render fine-grained medical images with high fidelity. Comprehensive experiments validate MedXChat's synergistic enhancement across all tasks. Our instruction data and model will be open-sourced.

Multimodal Large Language Models (MLLMs) have shown success in various general image processing tasks, yet their application in medical imaging is nascent, lacking tailored models.
This study investigates the potential of MLLMs in improving the understanding and generation of Chest X-Rays (CXRs). We introduce MedXChat, a unified framework facilitating seamless interactions between medical assistants and users for diverse CXR tasks, including text report generation, visual question-answering (VQA), and Text-to-CXR generation. Our MLLMs using natural language as the input breaks task boundaries, maximally simplifying medical professional training by allowing diverse tasks within a single environment.
For CXR understanding, we leverage powerful off-the-shelf visual encoders (\textit{e.g.}, ViT) and LLMs (\textit{e.g.}, mPLUG-Owl) to convert medical imagery into language-like features, and subsequently fine-tune our large pre-trained models for medical applications using a visual adapter network and a delta-tuning approach.
For CXR generation, we introduce an innovative synthesis approach that utilizes instruction-following capabilities within the Stable Diffusion (SD) architecture. This technique integrates smoothly with the existing model framework, requiring no extra parameters, thereby maintaining the SD's generative strength while also bestowing upon it the capacity to render fine-grained medical images with high fidelity.
Through comprehensive experiments, our model demonstrates exceptional cross-task adaptability, displaying adeptness across all three defined tasks. Our MedXChat model and the instruction dataset utilized in this research will be made publicly available to encourage further exploration in the field.
\end{abstract}

\section{Introduction}
\label{sec:intro}

Multi-modal tasks bridging image and text modalities play a pivotal role in advancing medical image analysis. Tasks such as medical report generation
~\cite{lee2023unified,thawkar2023xraygpt,lee2023llmcxr,vaswani2017attention,cornia2020meshed,wang2022medical}, medical visual question-answering (VQA)~\cite{xu2023elixr,thawkar2023xraygpt,lee2023llmcxr}, and \\medical visual grounding~\cite{lee2023llmcxr,lee2023unified} have gained increasing importance in providing comprehensive insights into complex medical data. Large Language Models (LLM), such as those based on transformer~\cite{vaswani2017attention} architectures, can play a wide role in promoting multimodal analysis due to their excellent capability in natural language processing (NLP) and reasoning.  Leveraging LLMs for multi-modal tasks offers the potential to enhance the understanding and interpretation of medical images through the fusion or transformation of textual and visual information.  Recent advancements in LLMs, exemplified by successes like Llama2\cite{touvron2023llama} and ChatGPT-4~\cite{openai2023gpt4}, highlight their effectiveness in various vision-and-language tasks ~\cite{koh2023generating,ge2023making,wu2023nextgpt}. These pre-trained LLMs are gaining traction in the medical domain for tasks like generating medical reports and answering questions~\cite{thawkar2023xraygpt}, yet there is a relative scarcity in text-to-medical-image generation and unified frameworks for handling diverse multimodal tasks in a single environment.  
Though such a unified large-scale multimodal data processing framework presents a significant opportunity, there is a noticeable absence for a multimodal processing framework for medical image and text.
To the best of our understanding, within the realm of conventional medical imaging, specifically chest X-rays (CXRs), there exist merely two comprehensive multimodal medical frameworks that facilitate both interpretation and creation. These are the non-LLM based method UniXGen~\cite{lee2023unified} model and the LLM-based method LLM-CXR~\cite{lee2023llmcxr}.
% Specifically,
% UniXGen employs the vector quantization (VQ)~\cite{esser2021taming}  approach 
% to perform both CXR interpretation and report generation by discretizing CXRs into distinct visual tokens as in sequence generation tasks. 
Specifically, UniXGen utilizes the vector quantization (VQ)~\cite{esser2021taming} technique to handle both the CXRs interpretation and generation of reports. 
% This method works by converting CXRs into discrete visual tokens, similar to the process used in sequence generation tasks.
% into sequence generation tasks. 
% While, LLM-CXR~\cite{lee2023llmcxr} employs VQ-GAN~\cite{esser2021taming}  to integrate image-text token space through the instruction-tuning technique that augments a text-only pre-trained LLM with the ability to process and understand medical images through vision-language integration. 
Meanwhile, the LLM-CXR again leverages VQ strategy to merge the image-text token space using the instruction-tuning approach. 
% This method enhances a pre-trained text-only LLM by equipping it with the capability to analyze and comprehend medical images through the integration of vision and language.
%
% Both models rely on the VQ technique to align textual and visual features, 
% Despite using VQ strategy can help to align textual and visual features, it inevitably brings a few additional limitations such as quantization errors and fixed codebook length, which will restrict the representation capacity. 
% and may hinder fine-grained CXR generation. It also limits the generalizability (e.g., when incorporating new datasets or expanding the codebook, it necessitates complete retraining, complicating the process).
Though the use of VQ strategy aids in aligning textual and visual features, it introduces several limitations, including quantization errors and a fixed codebook length.
% which can constrain the capacity for representation. This may also restrict generalizability, 
Also, this may restrict generalizability, \textit{e.g.}, incorporating new datasets or expanding the codebook requires full retraining, adding complexities. Therefore, there is a need for efficient and broadly applicable framework for processing large-scale multimodal medical data.
%

%First, they suffer from the quantization errors and the fixed length of the codebook for VQ limits the representation capacity. Given the subtle nuances within medical image, this constraint may prevent the generation of fine-grained CXR images. Moreover, the need for model and data updates—such as incorporating new datasets or expanding the codebook’s size or dimensions—necessitates a complete retraining of both the model and codebook, complicating the training process significantly.

To address aforementioned limitations and promote multimodal medical data processing, we introduce MedXChat, a comprehensive MLLM framework designed as a proficient medical assistant. This framework facilitates seamless interactions between medical data generation and understanding, enabling essential clinical capabilities such as image-to-report generation, intelligent question answering, and report-to-image synthesis. 
Specifically,
MedXChat is constructed on the foundation of instruction-tuning LLMs, addressing the image-text disparities from three perspectives. 
First, rather than relying on the local image patterns from discretely encoding such as VQ strategy as in ~\cite{lee2023llmcxr,lee2023unified}, we adopt visual features extracted and aligned by using CLIP encoder~\cite{radford2021learning}, which is sufficiently pre-trained to explicitly align image and text features using contrastive learning. Such strategy establishes a more intimate mapping between the image and text tokens, which is particularly beneficial for fine-grained medical data.  
Second, we employ close-sourced text-only ChatGPT-4~\cite{openai2023gpt4} API as the instructor to fine-tune our own LLM model and devise a way to construct instruction data by creating dialogues that refer to both medical reports and images. 
Third, we further address the image-text disparities in the text-to-image synthesis task by directly generating CXRs from input text prompt output by our fine-tuned LLM, avoiding the conversion from text tokens to image tokens as in conventional unified multimodal models~\cite{koh2023generating,ge2023making,wu2023nextgpt,lee2023llmcxr,lee2023unified}; 
this is achieved by leveraging the instruction-following capability inherent to the stable diffusion model~\cite{rombach2022high}. Our approach circumvents the full updates to the image generation component, enabling us to fine-tune the diffusion model portion without the need to retrain the entire language model. 

We conduct comprehensive experiments to validate the efficacy of our framework on CXR-to-Report, CXR-VQA and Text-to-CXR tasks respectively. We outperform most LLM-based methods~\cite{lee2023llmcxr,lee2023unified,thawkar2023xraygpt,wu2023towards,li2024llava} and non-LLM-based me-thods~\cite{chen2022cross,chen2020generating,cornia2020meshed} on those benchmarks. Moreover, to further validate whether our methods can contribute to clinical field, we invite a professional radiologist to assess 20 generated reports and 20 CXRs, utilizing UniXGen~\cite{lee2023unified} and LLM-CXR~\cite{lee2023llmcxr} for the evaluation. The outcomes again demonstrate that our methodology surpasses existing techniques in performance. Our main contributions are summarized as follows. 

% task was assessed using both traditional methods~\cite{wang2023metransformer,chen2020generating,chiang2023vicuna} and LLM-based methods~\cite{lee2023llmcxr,lee2023unified,thawkar2023xraygpt,li2024llava}. Subsequently, the CXR-VQA task underwent evaluation with LLM-based methods~\cite{xu2023elixr,thawkar2023xraygpt,wu2023towards,lee2023llmcxr}. The evaluation was completed with the Text-to-CXR task, again using LLM-based methods~\cite{lee2023unified,lee2023llmcxr}.  
\begin{itemize}
    \item We present MedXChat, 
    % a unified MLLM framework that facilitates the exchange of information between texts and CXR images within the medical domain.
    % a unified MLLM framework that facilitates both understanding and generation tasks for CXR images by exchanging information between texts and CXR images. It has demonstrated superiority across three key tasks: CXR-to-Report, CXR-VQA and Text-to-CXR tasks, compared with existing benchmark multimodal methods.
    a unified MLLM framework that supports both interpretive and generative functions for CXR images through the interchange of information between medical texts and CXRs images. It has demonstrated superiority across three key tasks: CXR-to-Report, CXR-VQA and Text-to-CXR tasks, compared with existing benchmark multimodal methods. 
    %
    % Our model has been comprehensively trained on the MIMIC dataset, encompassing both frontal and lateral views. It has demonstrated superiority across all three key tasks: CXR-to-Report, CXR-VQA and Text-to-CXR tasks, compared with the benchmark multimodal models.
    % We trained our model trained on the MIMIC dataset using the images in the frontal and lateral views. 
    %
    % \item In particular, we have pioneered a novel Text-to-CXR synthesis technique leveraging the instruction-following capabilities within the off-the-shelf Stable Diffusion method, which reserves the preservation of fine-grained feature generation essential for CXR imaging while harnessing the robust generative performance of the large pre-trained model.
    \item In particular, we have pioneered an innovative Text-to-CXR synthesis approach by utilizing the instruction-following abilities of the off-the-shelf Stable Diffusion method. This technique effectively maintains the creation of fine-grained features crucial for CXR imaging and takes advantage of the strong generative capabilities of the extensively pre-trained model.
    \item Our MedXChat also innovates on application side. 
 It uses natural language as the input and breaks task boundaries. This maximally simplifies medical professional training by allowing diverse tasks within a single environment, facilitating the transformation of AI to clinic utility. Also, our Text-to-CXR generation adds essential visual context to text, particularly for nuanced details, potentially benefiting radiologist education and clinical consultation.
    % \item We have developed a substantial amount of instruction data, specifically trained for Text-to-CXR generation. Additionally, we fine-tune SD model using ample medical data. All of these will be open-sourced.
    \item We have gathered a significant volume of instructional data from the MIMIC-CXR dataset, tailored specifically for Text-to-CXR generation. Furthermore, we have fine-tuned the Stable Diffusion (SD) method with a comprehensive collection of medical data. All these resources will be made available as open-source to facilitate the subsequent medical imaging generation research.

\end{itemize}

\begin{figure*}[t]
\begin{center}
\centerline{\includegraphics[width=1\linewidth]{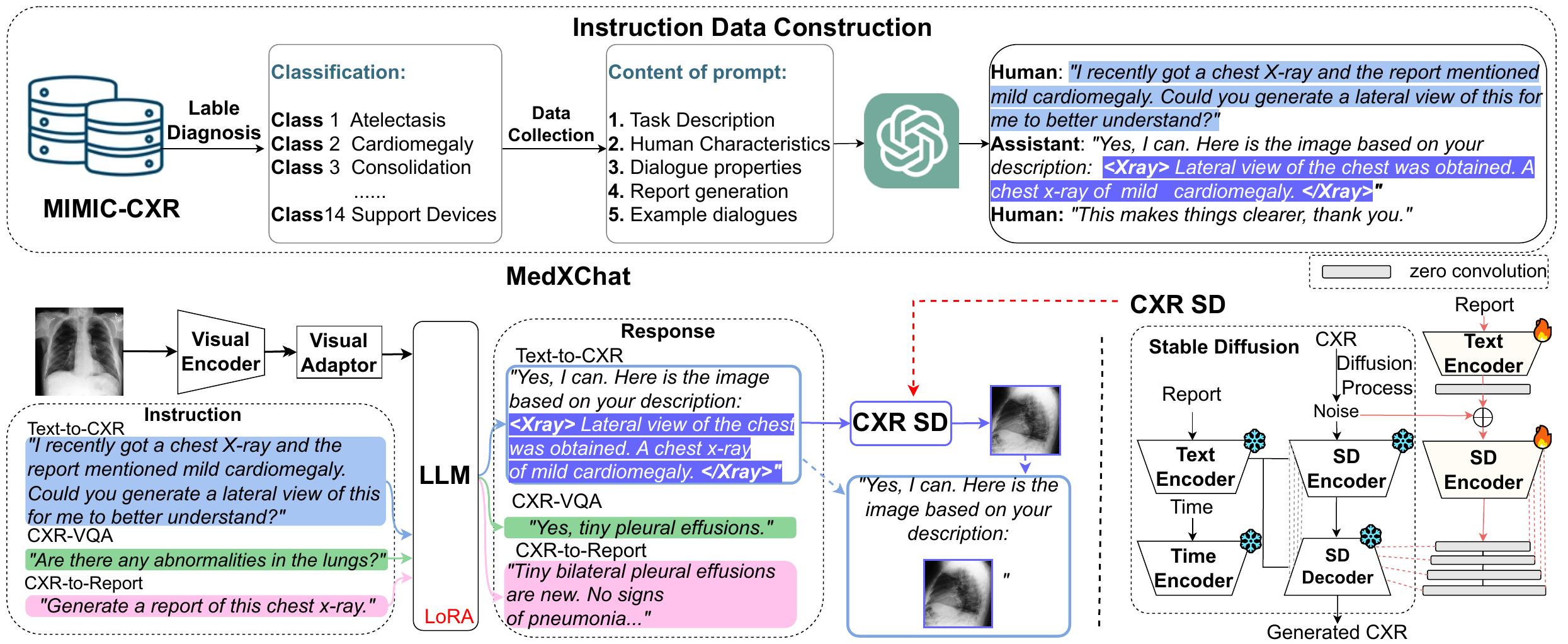}}
\end{center}
\vspace{-8mm}
\caption{Overview of our MedXChat framework, including a preparation stage (dashed boxes) for constructing instruction data (top row) and fine-tuning the Stable Diffusion model using CXR images (referred to as CXR-SD), followed by an instruction tuning stage (solid box) where our multimodal MedXChat is actively trained.}
\label{fig:framework}
%\vspace{-1em}
\end{figure*}
\section{Related Work}
\label{sec:related-work}
\textbf{Radiology report generation}
is rooted in image captioning but with specific challenges due to the fine-grained nature of medical images, the generation of long text, and the data distribution biased towards normal samples and observations. Existing methods in this field focus on addressing these challenges. For example, the works~\cite{wang2021self,you2021aligntransformer,wang2022automated} enhanced the alignment of image and text features to capture and describe fine-grained changes. The works~\cite{jing2017automatic,yin2019automatic,wang2021self} utilized hierarchical recurrent neural networks or advanced language model GPT-3~\cite{brown2020language} to improve long paragraph generation. To battle with data bias, many works~\cite{wang2022automated,jin2023promptmrg,jing2017automatic,tanida2023interactive,wang2022medical} trained additional disease classifiers to capture disease-related image features. More recent methods constructed knowledge graphs to guide report generation with disease prior knowledge~\cite{li2019knowledge,zhang2020radiology,liu2021exploring}, or contrasted disease images with normal samples~\cite{yang2022knowledge}. 
% [Please double check if there are LLM-based medical report generation methods recently]

\noindent\textbf{Medical Visual Question-answering (VQA)}
could be categorized into the dominant classification-based approaches~\cite{nguyen2019overcoming,eslami2021does,do2021multiple,finn2017model} and the generation-based approaches~\cite{ambati2018sequence,khare2021mmbert}. The field is advanced by integration with LLMs. 
% ChatCAD+~\cite{zhao2023chatcad+} is capable of handling medical images from diverse domains and leveraging up-to-date information from reputable medical websites to provide reliable medical advice. 
ELIXR~\cite{xu2023elixr} employed a language-aligned image encoder, integrated with or adapted to a fixed LLM, PaLM 2, to enhance VQA accuracy. 
RadFM~\cite{wu2023towards} an innovative radiology foundation model, by constructing MedMD, a vast medical multimodal dataset, and introducing an architecture for visually conditioned generative pre-training to evaluation benchmark encompassing VQA task.
Moreover, XrayGPT~\cite{thawkar2023xraygpt} aligned the medical visual encoder (MedClip)~\cite{wang2022medclip} with the fine-tuned LLM (Vicuna), showcasing remarkable capabilities in visual dialogue.

\noindent\textbf{Chest X-ray (CXR) Image Generation}
from radiology reports is very challenging and less researched despite the recent efforts~\cite{chuquicusma2018fool,baur1804melanogans,frid2018synthetic,wolleb2022diffusion,chambon2022adapting,chen2022exploiting} in simulating medical image data using deep generative models like variational autoencoders (VAE)~\cite{kingma2013auto}, generative adversarial networks ~\cite{goodfellow2020generative}, and diffusion models~\cite{ho2020denoising}. These simulation works did not make full use of the information contained in medical reports to generate details in CXR images. A more related work is~\cite{hou2023diversity}, which employed a domain-specific hierarchical text encoder to extract key medical concepts from reports, facilitating the generation of diverse and relevant X-ray images. It is noteworthy that none of the above systems could produce view-specific chest X-ray images or accommodate inputs from multiple perspectives, whereas our model could achieve.

\noindent\textbf{Multimodal Models}~~~
Text and image processing evolved independently until the Transformer model~\cite{vaswani2017attention}, ushering in an era of heightened correlation between the two fields. 
GILL~\cite{koh2023generating} translated text hidden representations into visual embedding space using unique image tokens. SEED~\cite{ge2023making} introduced a discrete image tokenizer for LLMs to see and draw simultaneously. In medical images, UniXGen~\cite{lee2023unified} generated bidirectional CXR and report, augmenting capabilities with special-view inputs. LLM-CXR~\cite{lee2023llmcxr} used transformer and VQ-GAN~\cite{esser2021taming}, instruction-tuning a text-pretrained LLM for enhanced alignment in understanding and generating CXR images. \underline{\textbf{Unlike}}  GILL and SEED, which extend LLM to text-to-image generation through a mapping network using special image tokens to bridge image-text disparities, we directly let LLM yield the text prompt allowing SD model to generate images, bypassing the potential token conversion gap. Also, compared with our MedXchat, UniXGen lacks VQA capabilities while LLM-CXR is incapable of handling multi-view input. Neither of them could generate lateral-view CXRs.

\section{Methodology}
\label{sec:method}
We propose a unified MLLM framework MedXChat, which possesses the capability to handle both multimodal medical data (\textit{e.g.}, image and text), while producing corresponding outputs. An overview of our framework is provided in Fig.~\ref{fig:framework} bottom right.
Specifically,
we select off-the-shelf LLM framework mPLUG-Owl~\cite{ye2023mplug} as our LLM due to its demonstrated effectiveness in numerous open-source projects focused on instructions-tuning language~\cite{taori2023stanford,chiang2023vicuna,peng2023instruction}. 
When optimizing this framework, the majority of the model weights, including those fundamental to the LLM and Stable Diffusion (SD), are kept frozen. For optimizing the main LLM, we use delta-tuning strategy, Low-Rank Adaptation (\textit{i.e.}, Lora)~\cite{hu2021lora}. 
The input to our model could be composed of a CXR  and the users' text instruction. The input CXR is processed by a pre-trained CLIP visual encoder ViT-L/14~\cite{radford2021learning} to extract text-aligned regional visual features to cater to the fine-grained nature of medical images and reports. These visual features are then combined with the users' instruction to form a passage of text fed into LLM. Consequently, our LLM makes responses based on the instructions. In particular, if our model parses the need for CXR generation, it will output prompts that can be used for the successive CXR synthesis. These prompts are then forwarded to our CXR-SD model, which fine-tunes the original SD on CXR images. Our training based on instruction tuning comprises a preparation stage to construct instruction data (Sec.~\ref{datadesign}) and fine-tune SD by CXR images (Sec.~\ref{cxr-sd}), and the actual instruction tuning stage for multimodal model training (Sec.~\ref{tasks} \&~\ref{Objective}).

\subsection{Instruction Data Construction}
\label{datadesign}
% \textcolor{red}{give a reason why you need to constuct the instruction}
% To construct instructions, we design conversations with enhanced human responsiveness using GPT-4~\cite{openai2023gpt4} as follows.
It is noteworthy that most current open-source LLMs~\cite{thawkar2023xraygpt,li2024llava} are incapable of handling image data while our unified model aims to possess image generation capabilities. Hence, to optimize our end-to-end MLLM to conduct conversations with patients in the manner of a clinician for text-to-CXR tasks. We need to obtain the instruction ground-truth so we can use such instruction for the fine-tuning process.
In this work,
we adopt ChatGPT-4~\cite{openai2023gpt4} and employ prompt engineering to generate a passage of text including a task description, human characteristics, dialogue properties, report generation, and example dialogues, as shown in the top of Fig.~\ref{fig:framework}.

Specifically, we provide five contents of prompts:
\textbf{1. Text Description} provides a general guideline for the response, \textit{e.g.}, \textit{``Please help construct three dialogues between a person and a helpful assistant. Each chest X-ray image is represented by $<Xray>$ DESCRIPTION $</Xray>$, where DESCRIPTION is a textual description or report of the X-ray image"}. \textbf{2. Human Characteristics} allows users to instruct the assistant on their individual need in generating chest X-ray images, \textit{e.g.}, whether the posteroanterior (PA) view or the lateral view is preferred. \textbf{3. Dialogue Properties} defines the properties of a good dialogue to ensure logical flow and restrict the content to only the pertinent parts of the report. \textbf{4. Report Generation} involves selecting a medical report from the MIMIC-CXR dataset and completing it with the view information. \textbf{5. Example Dialogues} provides two typical conversation examples within our framework. 

In this way, our LMM can then introduce special start and termination symbols $<Xray>$ and $</Xray>$ in the text-to-CXR generation section to signify that the intervening text corresponds to a text-represented image. 
For example, \textit{``the constructed dialogues must contain questions according to the following input chest x-ray image, which contains only the relevant part instead of the whole text between $<Xray>$ and $</Xray>$: $<Xray>$ \{content\} $</Xray>$"}. The position of \textit{``\{content\}"}  will be filled by the image's corresponding report augmented with different view information. We categorize the view of CXRs by training a transformer encoder upon all CXRs in the training set of MIMIC-CXR dataset to equip our model with the ability to differentiate between PA and lateral CXRs. Subsequently, we prepend the identified view information (\textit{i.e.}, \textit{``PA view of the chest was obtained"}), at the beginning of the relevant report. For the subsequent fine-tuning of the LLM and SD models, we utilize reports augmented with the various view information rather than solely relying on the original reports.

The passage of text is then input into ChatGPT-4 to produce three segments of medical dialogue. Within the MIMIC-CXR dataset, which contains paired CXRs and reports for 14 diagnostic categories, we select 200 reports from each category, adding up to 2800 reports in total. Each report is used to generate three corresponding dialogue segments by ChatGPT-4. Once the instruction data is constructed, all the generated dialogues are input back into our LLM, which is then fine-tuned using delta-tuning strategy LoRA to equip the model with dialogue analysis capabilities.

By fine-tuning the pre-trained LLM, we can then produce accurate dialogue content. Once there is a need to generate a CXR that corresponds to a report, our LLM will be correspondingly trained to identify the special symbols $<Xray>$ and $</Xray>$, where the content in between is then extracted as the prompt fed into our CXR-SD for text-to-image generation. To allow users to generate CXR images from user-specified views, we further fine-tune the original SD model with CXRs and our view-augmented reports, enabling the precise generation of CXR images that match the specified views.

\begin{figure*}[h]
\begin{center}
\centerline{\includegraphics[width=1\linewidth]{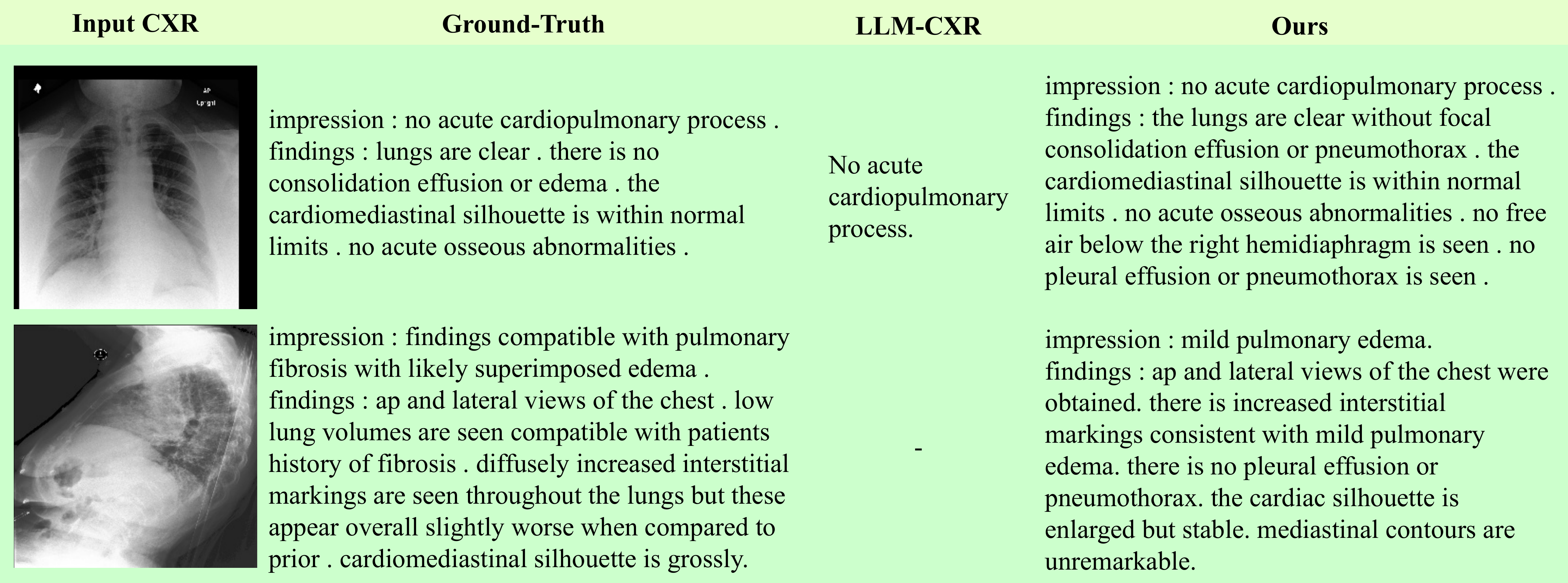}}
\end{center}
\vspace{-8mm}
\caption{Visual examples for the CXR-to-Report Task}
\label{fig:cxr-to-report}
%\vspace{-1em}
\end{figure*}
\vspace{-15mm}
\begin{figure*}[!h]
\begin{center}
\centerline{\includegraphics[width=1\linewidth]{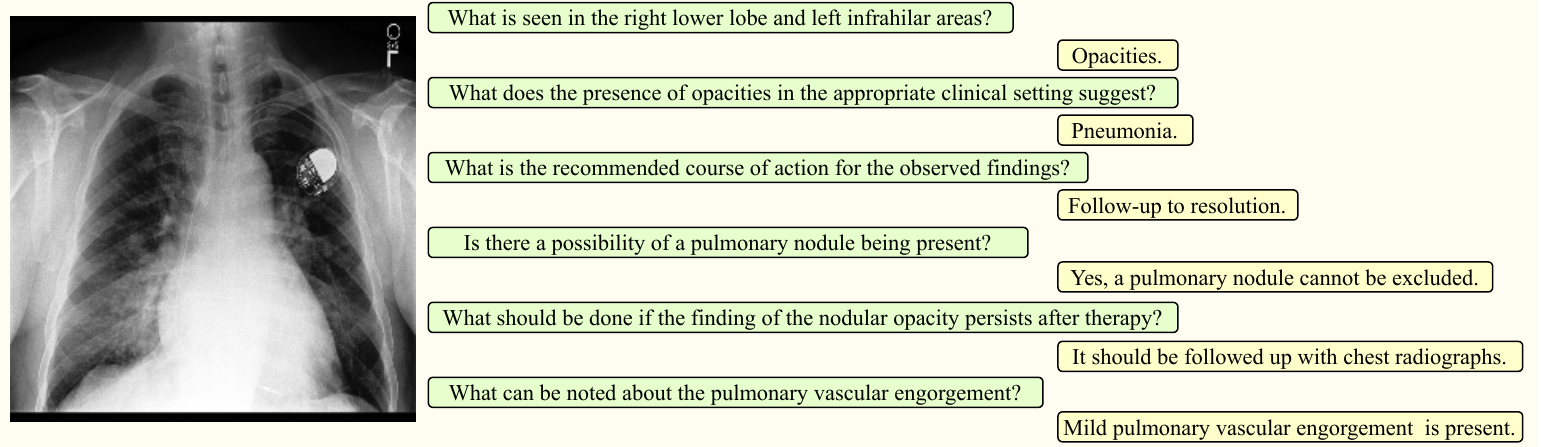}}
\end{center}
\vspace{-8mm}
\caption{Visual examples for the CXR-VQA Task}
\label{fig:cxr-vqa}
%\vspace{-1em}
\end{figure*}

\subsection{CXR SD}
\label{cxr-sd}
To generate CXR from a given clinical test,
we utilize a pre-trained SD model~\cite{rombach2022high} generation network. Similar to recent popular fine-tuning strategy~\cite{zhang2023adding}, we fine-tune our SD based on MIMIC-CXR dataset using the zero-convolution module as shown in Fig.~\ref{fig:framework} bottom left. 

For understanding the data distribution $p(z)$, SD will progressively denoise variables that follow a normal distribution using an encoder-decoder style network, which is equivalent to mastering the reverse process of a predetermined Markov chain of time-step length $T$. In each time-step $t$, we first encode the text features using a time encoder ${E}_t$ employing positional encoding techniques. 

In the beginning, with a sample CXR $z_0$, SD incrementally adds noise to $z_0$ via a forward diffusion process denoted as $q$:
\begin{equation}
q\left(z_t \mid z_0\right)=\mathcal{N}\left(z_t ; \alpha_t z_0, \sigma_t^2 \mathbb{I}\right).
\end{equation} 

After $T$ timesteps, the initial CXR will be transformed into a pure noise $z_t$, and then sent into a encoder-decoder 
 architecture (\textit{i.e.}, U-Net~\cite{ronneberger2015u}) %featured by an encoder ${E}_{sd}$ and a decoder ${D}_{sd}$ with skip connections. 
including 4 down-sampling blocks and 4 up-sampling blocks, respectively. The remaining core blocks each consist of four ResNet layers and two Vision Transformers (ViTs), with each ViT incorporating multiple cross-attention and self-attention mechanisms~\cite{vaswani2017attention}. 
%\begin{equation}
%   \mathrm{Att}({\mathbf Q}, {\mathbf K}, {\mathbf V}) = \mathrm{softmax}\left(\frac{{\mathbf Q}{\mathbf K}^{T}}{\sqrt{d_k}}\right){\mathbf V}.
%\label{equ:attention}
%\end{equation} 
%
Then, SD takes text prompts that undergo encoding via using the CLIP text encoder ${E}_{clip}$~\cite{radford2021learning} and then merge with the noise by forecasting and implementing the reverse process of noise addition:
\begin{equation}
\begin{aligned} p_\theta\left(z_{t-1} \mid z_t\right) & :=q\left(z_{t-1} \mid z_t, \epsilon_\theta\left(z_t, t\right)\right) \\ & =\mathcal{N}\left(z_{t-1} ; \mu_\theta\left(z_t, t\right), \sigma_{t \mid t-1}^2 \frac{\sigma_{t-1}^2}{\sigma_t^2} \mathbb{I}\right), \end{aligned}
\end{equation} where $\epsilon_\theta$ is the mapping function of a U-Net model predicting noise.
This method systematically lessens the noise, ultimately generating the corresponding CXR.

We have adopted the pre-trained SD based on a large dataset, which already yields impressive outcomes in text-to-image generation. Nonetheless, when it comes to medical datasets, fine-tuning remains essential to guarantee the quality of the generated results. During our initialization stage, to preserve the inherent capabilities of the original SD without compromising its performance, we introduce a module that only trains SD encoder ${E}_{sd}$ and several $1 \times 1$ zero-convolution layers, where their weights were initialized as zero values adding no impact to the deep features. Then, they will gradually increase from zero to optimized parameters during the subsequent fine-tuning process. Since we are only plugging the proposed module into the pre-trained model, this training is as fast as fine-tuning diffusion models compared to training from scratch.

During our fine-tuning stage, the report is initially processed through a text encoder, succeeded by a zero-convolution layer. Subsequently, it is concatenated with noise $z_t$ as part of the denoising process which is achieved by passing the data through four down-sampling blocks of the trainable SD encoder, with each block aligning with the corresponding four up-sampling blocks of the SD encoder via a zero convolution layer. Ultimately, these results are integrated with the outcomes of the fixed SD encoder from the pre-trained model. By adopting this approach, we only focus on training our specialized model for rapid fine-tuning rather than the entire SD model. This strategy maintains the efficacy of the pre-trained model while also enabling the generation of fine-grained CXR.
% \vspace{-10mm}
%
\begin{figure*}[h]
\begin{center}
\centerline{\includegraphics[width=1\linewidth]{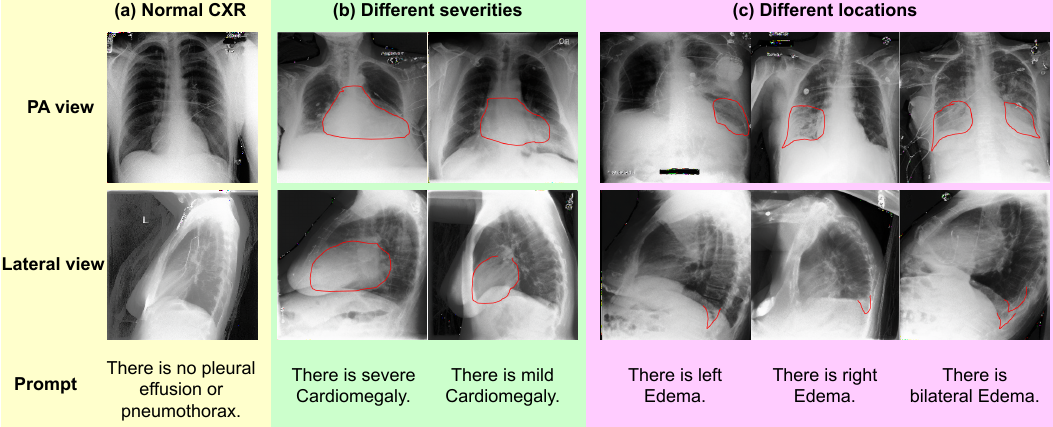}}
\end{center}
\vspace{-8mm}
\caption{Visual examples showcasing our Text-to-CXR generation. The delineated image regions correspond to the users' prompts, expertly identified by a trained Radiologist.}
\label{fig:text-to-CXR}
%\vspace{-1em}
\vspace{-6mm}
\end{figure*}
\subsection{Multi Model Tasks}
\label{tasks}
% instruction tuning
The proposed framework is a unified framework shown in Fig.~\ref{fig:framework} bottom right, integrating three distinct tasks: CXR-to-Report, CXR-VQA and Text-to-CXR, elaborated below. 
%In this section, we will delve into the design specifics for each individual task.

\noindent\textbf{CXR-to-Report}~~~In this task, a pre-trained CLIP visual encoder ViT-L/14~\cite{radford2021learning} is used to process the CXR image, providing text-aligned visual features. These visual features are then paired with a corresponding user's instruction, \textit{e.g.}, \textit{"Generate a report of this chest x-ray"}, and input into the LLM. Based on this instruction, LLM produces a concise report according to the information extracted from the CXR. 

\noindent\textbf{CXR-VQA}~~~In this task, our model inputs a set of three conditions into the LLM: the CXR image, the accompanying medical report, and a question from the user like \textit{"Is there cardiomegaly?"}. The LLM evaluates these inputs and responds with an answer, for instance, \textit{"No, cardiomegaly is not present." }

\noindent\textbf{Text-to-CXR}~~~In this task, the dialogues generated from Sec.~\ref{datadesign} are sent to LLM for training. An example of the prompt input to LLM is as follows.

\textit{"Human: Hi there, I recently had a chest X-ray and the report mentioned "right lower lobe consolidation". I'm not sure what that means. Could you generate a lateral view of a chest X-ray with that condition for me?"}

The output of LLM for this example is like:

\textit{Assistant: Absolutely. Consolidation in the right lower lobe of your lungs typically indicates an infection, such as pneumonia. Here is an image based on the description you provided: $<Xray>$ Lateral view of the chest was obtained. New right lower lobe consolidation is consistent with pneumonia. $</Xray>$.}

The LLM processes this textual description and extracts the content between $<Xray>$ and $</Xray>$ as the prompt for CXR synthesis, which is further passed to our CXR-SD for CXR generation. Our model is able to generate a CXR image with meticulous details, enabling the LLM to respond to the user's request with a visual representation.

\subsection{Training Objectives}
\label{Objective}
Our training objective encompasses two components: LLM at the instruction tuning stage and CXR-SD at the preparation stage. LLM's objective is to produce comprehensive target paragraphs consisting of an explanation, input, and an autoregressive response, akin to the instruction-following fine-tuning paradi-gm~\cite{taori2023stanford} used in ChatGPT pre-training~\cite{radford2018improving,brown2020language}. The loss function is specifically calculated based on the tokens generated in response to the prompt, aligning with the key response mechanism of the model. Moreover, the training objective of CXR-SD follows that of the SD model, focusing on denoising~\cite{rombach2022high}: it employs textual prompts and temporal training steps to forecast the noise that has been introduced to the distorted images. Through this process, the model incrementally diminishes the noise, thereby progressively restoring the clarity of the image with each step.

\noindent\textbf{Large Language Model.}
%Leveraging the approaches outlined in Sec.~\ref{datadesign}, we've developed an extensive collection of dialogues $\left(\mathbf{X}_{\mathrm{q}}^1, \mathbf{X}_{\mathrm{a}}^1, \cdots, \mathbf{X}_{\mathrm{q}}^T, \mathbf{X}_{\mathrm{a}}^T\right)$ related to the reports from the MIMIC dataset utilizing ChatGPT-4, where $\mathbf{X}_{\mathrm{q}}$ is the question, $\mathbf{X}_{\mathrm{a}}$ is the answer and T is the total number of dialogues. These dialogues underwent a fine-tuning process with LoRA~\cite{hu2021lora} through the loss $L_a$:
With the dialogues $\left(\mathbf{X}_{\mathrm{q}}^1, \mathbf{X}_{\mathrm{a}}^1, \cdots, \mathbf{X}_{\mathrm{q}}^T, \mathbf{X}_{\mathrm{a}}^T\right)$ obtained utilizing ChatGPT-4 in Sec.~\ref{datadesign}, where $\mathbf{X}_{\mathrm{q}}$ is the question, $\mathbf{X}_{\mathrm{a}}$ is the answer, and T is the total number of dialogues, we fine-tune LLM with LoRA~\cite{hu2021lora} through minimizing the loss $L_a$:
\begin{equation}
L_a=\sum_{i=1}^{L}-\log p\left(\mathbf{X}_{\mathrm{a}}^i \mid \mathbf{X}_{\mathrm{q}}^t,<i, \mathbf{X}_{\mathrm{a}}^t,<i\right)
\end{equation}
for a sequence of length L.
As a result, the enhanced LLM is now adept at crafting intelligent and contextually relevant medical responses. 

\noindent\textbf{Stable Diffusion.}
Given an initial image $z_0$, the image diffusion process incrementally introduces noise, resulting in a sequence of increasingly noisy images $z_t$, where $t$ denotes the timestep of noise addition. Under a set of conditions, comprising the timestep 
$t$ and textual prompts $c_t$, a network $\epsilon_\theta$ is trained to predict the noise pattern that has been added to the noisy image $z_t$, enabling the reverse process of noise reduction to reconstruct the original image. The overall learning objective for the entire diffusion model is to minimize the loss $\mathcal{L}_{SD}$:
\begin{equation}
\left.\mathcal{L}_{SD}=\mathbb{E}_{\boldsymbol{z}_0, \boldsymbol{t}, \boldsymbol{c}_t, \epsilon \sim \mathcal{N}(0,1)}\left[\| \epsilon-\epsilon_\theta\left(\boldsymbol{z}_t, \boldsymbol{t}, \boldsymbol{c}_t \right)\right) \|_2^2\right].
\end{equation}

\section{Experiments}
\label{sec:experiments}
We benchmark our model against key state-of-the-art (SOTA) models for CXR-related multimodal generation using LLMs. For CXR-to-Report, our model's performance is mainly benchmarked against UniXGen~\cite{lee2023unified}, XrayGPT~\cite{thawkar2023xraygpt}, and LLM-CXR~\cite{lee2023llmcxr}. For CXR-VQA, we compare with XrayGPT~\cite{thawkar2023xraygpt}, ELIXR ~\cite{xu2023elixr}, and LLM-CXR~\cite{lee2023llmcxr}. For Text-to-CXR, we evaluate our model against UniXGen~\cite{lee2023unified}  and LLM-CXR~\cite{lee2023llmcxr}. It is highlighted that LLM-CXR and our model are uniquely equipped to handle all of these diverse tasks using a single integrated model. Whereas, XrayGPT~\cite{thawkar2023xraygpt} cannot handle Text-to-CXR generation and UniXGen cannot handle CXR-VQA task.

\subsection{Dataset}
\noindent\textbf{MIMIC-CXR}~\cite{johnson2019mimic} stands as the largest publicly available dataset encompassing chest radiographs along with free-text reports. It covers 14 categories related to lung disease diagnosis. Part of the CXR images are obtained in both PA and lateral views. In our research, we followed the official dataset division guidelines of MIMIC-CXR to maintain consistency and enable fair comparisons. Consequently, our training dataset comprises 270,790 samples of image-report pairs, accompanied by 2,130 and 3,858 samples for validation and testing, respectively. %More details about the dataset could be found in the supplement.

\begin{table*}[h]
\centering
\caption{CXR-to-Report: AUROC and F1. \dag~marks quoted results from~\cite{lee2023llmcxr}.}\label{Table:AUROC&F1}
\scalebox{0.9}{\begin{tabular}{l||ccc|cccccc}
\hline
AUROC$\uparrow$           & Micro & Macro & Weighted & NoF. & Pmtx. & Edem. & PEff. & Csdn./Pna. & LLsn.  \\
\hline\hline
UniXGen-256\dag~~\cite{lee2023unified} & 0.577 & 0.533 & 0.541 & 0.564 & 0.530 & 0.542 & 0.533 & 0.516 & 0.513   \\
XrayGPT\dag~~\cite{thawkar2023xraygpt} & 0.595 & 0.552 & 0.576 & 0.592 & 0.511 & 0.590 & 0.595 & 0.515 & 0.511     \\
LLM-CXR\dag~~\cite{lee2023llmcxr} & 0.654 & 0.586 & 0.628 & \textbf{0.698} & 0.532 & 0.612 & 0.635 & \textbf{0.540} & 0.501    \\
MedXChat (Ours)& \textbf{0.672}    & \textbf{0.599}     & \textbf{0.666} & 0.583 & \textbf{0.549} & \textbf{0.633} & \textbf{0.783} & 0.530     & \textbf{0.519}               \\
\hline
\end{tabular}}
\scalebox{0.9}{\begin{tabular}{l||ccc|cccccc}
\hline
F1$\uparrow$           & Micro & Macro & Weighted & NoF. & Pmtx. & Edem. & PEff. & Csdn./Pna. & LLsn.  \\
\hline\hline
UniXGen-256\dag~~\cite{lee2023unified}   & 0.281 & 0.187 & 0.256 & 0.411 & 0.083 & 0.226 & 0.215 & 0.132 & 0.055  \\
XrayGPT\dag~~\cite{thawkar2023xraygpt} & 0.314 & 0.227 & 0.320 & 0.371 & 0.049 & 0.333 & 0.404 & 0.143 & \textbf{0.058}    \\
LLM-CXR\dag~~\cite{lee2023llmcxr} & 0.414 & 0.283 & 0.408 & \textbf{0.562} & 0.083 & 0.370 & 0.455 & \textbf{0.198} & 0.030     \\
MedXChat (Ours) & \textbf{0.420} & \textbf{0.292}  & \textbf{0.436} & 0.318 & \textbf{0.092} & \textbf{0.398} & \textbf{0.718} & 0.177   & 0.049               \\

\hline
\end{tabular}}
\end{table*}
\begin{table*}[!h]
\centering
\caption{CXR-to-Report: NLP Metrics. \dag~marks  quoted results from respective papers.}\label{Table:bleu}
%\vspace{-2mm}
\begin{tabular}{l||ccccccc}
\hline
Methods           & BLEU-1 & BLEU-2 & BLEU-3 & BLEU-4 & ROUGE & METEOR & CIDEr  \\
\hline\hline
% XrayGPT & -    & -     & - & - & - & - & -                \\
Show-Tell~\cite{vinyals2015show} & 0.308 & 0.190 & 0.125 & 0.088 & 0.256 & 0.122 & 0.096 \\
Att2in~\cite{rennie2017self} & 0.314 & 0.198 & 0.133 & 0.095 & 0.264 & 0.122 & 0.106 \\
AdaAtt~\cite{lu2017knowing} & 0.314 & 0.198 & 0.132 & 0.094 & 0.267 & 0.128 & 0.131 \\
Transformer~\cite{vaswani2017attention} & 0.316 & 0.199 & 0.140 & 0.092 & 0.267 & 0.129 & 0.134 \\
M2transformer~\cite{cornia2020meshed} & 0.332 & 0.210 & 0.142 & 0.101 & 0.264 & 0.134 & 0.142  \\
\hdashline
R2Gen\dag~\cite{chen2020generating} & 0.353 & 0.218 & 0.145 & 0.103 & 0.277 & 0.142 & 0.141  \\
PPKED\dag~\cite{liu2021exploring} & 0.36 & 0.224 & 0.149 & 0.106 & 0.284 & 0.149 & 0.237  \\
R2GenCMN\dag~\cite{chen2022cross} & 0.353 & 0.218 & 0.148 & 0.106 & 0.278 & 0.142 & 0.143  \\
GSK\dag~\cite{yang2022knowledge} & 0.363 & 0.228 & 0.156 & \textbf{0.115} & 0.284 & - & 0.203  \\
% MSAT~\cite{wang2022medical} & 0.373 & 0.235 & 0.162 & 0.120 & 0.282 & 0.143 & 0.299    \\
% METransformer~\cite{wang2023metransformer} & \textbf{0.386} & \textbf{0.250} & \textbf{0.169} & \textbf{0.124} & \textbf{0.291}  & \textbf{0.152} & \textbf{0.362} \\
\hdashline
LLM-CXR~\cite{lee2023llmcxr} & 0.196 & 0.095  & 0.054 & 0.033 & 0.245 & 0.081 & \textbf{0.445}               \\
LLaVA-Med~\cite{li2024llava} & 0.232 & 0.086 & 0.027 & 0.009 & 0.168 & 0.082 & 0.015 \\
UniXGen-256~\cite{lee2023unified}   & 0.365    & 0.227     & 0.147 & 0.101 & \textbf{0.294} & \textbf{0.156} & 0.138             \\
\hline
MedXChat (Ours) & \textbf{0.367}    & \textbf{0.235}     & \textbf{0.158} & 0.111 & 0.264 & 0.135 & 0.175              \\
\hline
\end{tabular}
\end{table*}
% \vspace{-10mm}

\subsection{Implementation}
\noindent\textbf{Downstream Classification.}~~~
%To evaluate the performance of our Text-to-CXR generation in capturing clinically useful information, 
To evaluate the clinical utility of our Text-to-CXR generation,
we apply our generated CXRs to classifying the 14 diagnostic categories in MIMIC-CXR. We adopt a classification model proposed in~\cite{he2015deep}, which gains high classification accuracy from its increased model depth via a residual learning framework, train it by all CXR images in MIMIC-CXR training set, and test it on the generated CXR images produced by different models.
%To test the useful clinic information of our text-to-CXR generated images, we classified all the training sets of the MIMIC dataset into 14 official categories for targeted training, and then categorized the generated test results.
\label{downstream}

\vspace{-3mm}
\subsection{Evaluation Metrics}
\noindent\textbf{CXR-to-Report.}
We include both NLP standard evaluation metrics such as BLEU-4~\cite{papineni2002bleu}, METEOR~\cite{banerjee2005meteor}, ROUGE-L~\cite{lin2004rouge}, and CIDEr~\cite{vedantam2015cider} to evaluate the quality of the generated diagnostic reports, and the AUROC and F1 score metrics to assess the accuracy of disease classification based on generated reports. As per~\cite{lee2023llmcxr}, we report AUROC and F1 in three forms: Micro, Macro, and Weighted, derived from the frequency deviation of categories, with each category being treated as equally important and weighted average, respectively. They are calculated based on six diagnostic categories: No Findings (NoF.), Pneumothorax (Pmtx.), Edema (Edem.), Pleural Effusion (PEff.), Consolidation or Pneumonia (Csdn./Pna.), and Lung lesion (LLsn.).

\noindent\textbf{CXR-VQA.}
We follow the VQA performance evaluation framework from ELIX-R~\cite{xu2023elixr} to enable the comparison with other LLM-based methods\footnote{This choice is necessitated by the unavailability of CXR-VQA results for other LLM-based methods}. Specifically, from the MIMIC-CXR dataset, we randomly choose eight samples from each of the six distinct diagnostic categories. We either inquiry about specific lesions or query the presence, location, and severity of the findings in each CXR image.

\begin{table}[!h]
\centering
\vspace{-5mm}
\caption{CXR-VQA: Accuracy by topic. \dag~marks quoted results from~\cite{lee2023llmcxr}.  ``SST" stands for Size, Severity, Type.} \label{Table:TopicAccuarcy}
\scalebox{0.9}{
    \begin{tabular}{l||c|ccc}
    \hline
    Accuracy$\uparrow$  & All & Presence & Location & SST  \\
    \hline\hline
    ELIXR\dag~~\cite{xu2023elixr}   & 54.8\% & 64.5\% & 41.0\% & 25.0\%             \\
    XrayGPT\dag~~\cite{thawkar2023xraygpt} & 25.2\% & 27.4\% & 21.9\% & 20.3\%          \\
    RadFM\dag~~\cite{wu2023towards} & 32.7\% & 34.5\% & 31.3\% & 20.8\%    \\
    LLM-CXR\dag~~\cite{lee2023llmcxr} & 44.8\% & 41.3\% & 50.0\% & 62.5\%    \\
    LLaVA-Med~\cite{li2024llava} & 53.1\% & 53.8\% & 51.0\% & 56.3\%       \\
    MedXChat (Ours) & \textbf{61.2\%} & \textbf{61.5\%}     & \textbf{56.3\%} & \textbf{68.8\%}             \\
    \hline
    \end{tabular}
}
\vspace{-10mm}
\end{table}
\begin{table*}[!h]
\centering
\caption{CXR-VQA: Accuracy by Diagnosis. %Cnsl.: Consolidation, Pna.: Pneumonia, Les.: Lung lesion (nodule or mass), NoF.: No Findings (i.e., normal CXR), Pmtx.: Pneumothorax. 
\dag~marks quoted results from~\cite{lee2023llmcxr}. %And ELIXR~\cite{xu2023elixr} does not report its VQA accuracy by label diagnosis.
}\label{Table:Label DiagnosisAccuarcy}
\begin{tabular}{l||c|cccccc}
\hline
Accuracy$\uparrow$  & All  & NoF. & Pmtx. & Edem. & PEff. & Csdn./Pna.  & LLsn.\\
\hline\hline
XrayGPT\dag~~\cite{thawkar2023xraygpt} & 25.2\% & 42.5\% & 18.8\% & 26.3\%  & 20.0\% & 25.0\% & 17.2\%\\
RadFM\dag~~\cite{wu2023towards} & 32.7\% & 61.3\% & 23.8\% & 31.3\%  & 26.3\% & 34.4\%  & 40.6\%     \\
LLM-CXR\dag~~\cite{lee2023llmcxr} & 44.8\% & \textbf{71.3\%} & 22.5\% & 53.8\%  & 53.8\% & 39.1\%  & \textbf{50.0\%}     \\
LLaVA-Med~\cite{li2024llava} & 53.1\% & 37.5\% & 54.2\% & 58.3\% & 60.4\% & \textbf{56.3\%} & 37.5\% \\
MedXChat (Ours) & \textbf{61.2\%} & 50.0\% & \textbf{58.3\%} & \textbf{72.9\%}  & \textbf{79.2\%} & 50.0\%  & 37.5\% \\
\hline
\end{tabular}
\end{table*}

\noindent\textbf{Text-to-CXR.}
We employ the Frechet Inception Distance (FID), a widely recognized metric for image generation quality assessment, in conjunction with a downstream classification task outlined in Sec.~\ref{downstream}, to evaluate the quality of the images produced by our model.

\subsection{Results}
\noindent\textbf{CXR-to-Report.}
The results of AUROC and F1 are compared in Table~\ref{Table:AUROC&F1}, which are critical metrics to reflect the clinic efficacy of the generated reports. The results of the comparing methods are directly quoted from the paper of LLM-CXR~\cite{lee2023llmcxr}. As seen, our model produces the highest overall scores in terms of both F1 and AUROC among the compared methods. It is also the best/second performer across all diagnostic categories except ``No Fundings". As also observed, the non-LLM-based UniXGen~\cite{lee2023unified} underperforms when compared with LLM-based methods XrayGPT~\cite{thawkar2023xraygpt} and LLM-CXR~\cite{lee2023llmcxr}, showing the potential of LLMs in generating clinic-relevant contents. 

NLP metrics are given in Table~\ref{Table:bleu}. 
%We leverage the LLM-CXR~\cite{lee2023llmcxr} API to obtain its generated reports. 
In addition to large multimodal generation methods, we also include conventional methods exclusively for text generation. As seen, our model's BLEU scores (BLEU-1 to BLEU-4) surpass nearly all methods, indicating a substantial overlap of n-grams in our generation and the ground truth. Moreover, compared with LLM-CXR, the sole method capable of performing all three tasks within a single integrated framework like ours, our model exhibits indisputable superiority across all NLP metrics except CIDEr. However, the notably high CIDEr of LLM-CXR might be linked to a distinct evaluation protocol focusing on the impression section rather than the complete report
%However, the notably high CIDEr of LLM-CXR might be linked to a distinctive evaluation: the ground truth provided by LLM-CXR comprises only the impression section of a report, which is just a one-sentence summary instead of the complete report
\footnote{CXR-to-Report performance evaluation of LLM-CXR using complete reports yields lower results than those reported in Table~\ref{Table:bleu}}. Ours also outperforms UniXGen~\cite{lee2023unified} in all BLEU scores and CIDEr. Notably, our model excels in generating clinically consistent reports as already seen in Table~\ref{Table:AUROC&F1}.
%More importantly, it is again worth mentioning that our model wins both methods in generating clinically consistent reports as already seen in Table~\ref{Table:AUROC&F1}. 
Compared to conventional methods, ours leads in BLEU-1 to BLEU-3, with only a slight lag behind GSK in BLEU-4. Our performance is average in ROUGE, yet superior to a majority of models in METEOR and CIDEr.

Moreover, a qualitative comparison is presented in Fig.~\ref{fig:cxr-to-report}. As evident, LLM-CXR~\cite{lee2023llmcxr} generates only front-view reports containing a brief impression session, while failing to handle lateral view CXRs. Conversely, our generated reports not only offer impressions but also furnish detailed findings on specific diagnoses, ensuring close alignment with the content of the original report.

\noindent\textbf{CXR-VQA.}~~Table~\ref{Table:TopicAccuarcy} and Table~\ref{Table:Label DiagnosisAccuarcy} present VQA accuracies for various multimodal LLMs. In Table~\ref{Table:TopicAccuarcy}, the answers are marked based on the accuracies of predicting the presence, location, and SST (Size, Severity, Type) of the lesions or findings. Our model achieves the highest overall accuracy, excelling in all three categories of prediction. In Table~\ref{Table:Label DiagnosisAccuarcy}, the answers are marked based on the accuracies of predicting the diagnostic labels. Again, our model stands out as the best performer, showing superior performance for Edema, Pleural Effusion, and Pneumothorax. An example of our VQA generation is illustrated in Fig.~\ref{fig:cxr-vqa}, showcasing the model's proficiency in answering questions regarding the categories and severities of the findings.

\begin{table}[h]
    \centering
    \begin{minipage}{0.45\linewidth}
        \centering
        \caption{Text-to-CXR: FID and Classification accuracy}\label{Table:FID}
        \scalebox{0.88}{
            \begin{tabular}{l||cc}
            \hline
            Method   & FID$\downarrow$  & Classification $\uparrow$\\
            \hline\hline
            UniXGen~\cite{lee2023unified} & 106.17   & 67.2\%           \\
            LLM-CXR~\cite{lee2023llmcxr} & 73.29    & 68.6\%          \\
            MedXChat (Ours) & \textbf{43.46}     & \textbf{71.5\%}      \\
            \hline
            \end{tabular}
        }
    \end{minipage}
    \hfill
    \begin{minipage}{0.45\linewidth}
        \centering
        \caption{Radiologist evaluation on 40 randomly selected generation cases}\label{Table:Rad-eva}
        \scalebox{0.88}{
            \begin{tabular}{l||cc}
            \hline
            \#prefer/total & Reports  & CXRs\\
            \hline\hline
            UniXGen~\cite{lee2023unified} & 1/20   & 7/20           \\
            LLM-CXR~\cite{lee2023llmcxr} & 0/20    & 0/20          \\
            MedXChat (Ours) & \textbf{19/20}     & \textbf{13/20}      \\
            \hline
            \end{tabular}
        }
    \end{minipage}
\end{table}

% \begin{table}[h]
% \centering
% \caption{Text-to-CXR: FID and Classification accuracy}\label{Table:FID}
% \scalebox{0.9}{
%     \begin{tabular}{l||cc}
%     \hline
%     Method   & FID$\downarrow$  & Classification $\uparrow$\\
%     \hline\hline
%     UniXGen~\cite{lee2023unified} & 106.17   & 67.2\%           \\
%     LLM-CXR~\cite{lee2023llmcxr} & 73.29    & 68.6\%          \\
%     MedXChat (Ours) & \textbf{43.46}     & \textbf{71.5\%}      \\
%     \hline
%     \end{tabular}
% }
% \end{table}

\noindent\textbf{Text-to-CXR.}
Table~\ref{Table:FID} presents the quantitative results of Text-to-CXR generation. Notably, our model yields a significantly lower FID score, indicating a much higher similarity between the distributions of real CXR images and our generated images. Crucially, to access the clinical utility of the generated CXR images, we train a classifier using authentic CXR images to classify the 14 diagnostic categories in MIMIC-CXR,  as described in Sec.~\ref{downstream}. The results show that our generated CXRs contribute to higher classification accuracies, suggesting a superior fidelity in capturing clinically informative patterns. An interesting observation is the non-LLM-based UniXGen~\cite{lee2023unified} exhibits inferiority in capturing clinical information compared to LLM-based methods, reinforcing our findings from the CXR-to-Report task. 

Visual examples in Fig.\ref{fig:text-to-CXR} showcase our model's reliable generation of abnormalities, aligning with specified clinical conditions, severities, locations, and views.
Beyond superior quantitative results, a significant advantage of our model over LLM-CXR~\cite{lee2023llmcxr} and UniXGen~\cite{lee2023unified} lies in our ability to generate high-quality lateral views, whereas the other two are limited in PA views only. Our model excels in capturing intricate visual patterns, as seen in bilateral edema (Fig.~\ref{fig:text-to-CXR} rightmost), where fine details of clinic conditions in the lateral view are successfully reproduced.

\noindent\textbf{Radiologist Evaluation.}
To further validate the performance of the three tasks, we facilitated a detailed comparative evaluation by a radiologist, focusing on the CXR-to-Report and Text-to-CXR tasks, as exemplified in Fig.\ref{Table:Rad-eva}. Acknowledging the impracticality of exhaustive manual evaluation due to the MIMIC-CXR dataset's extensive size, we randomly selected a representative subset of 40 generative cases for comparison. This subset included results generated by us, UniXGen~\cite{lee2023unified}, and LLM-CXR~\cite{lee2023llmcxr}, containing 20 reports and 20 frontal CXRs. The CXRs only featured the frontal view because UniXGen and LLM-CXR are capable of generating only frontal CXRs. The evaluations predominantly favored the reports and CXRs generated by our model, with only a minority preferring the outputs from UniXGen. This trend underscores our approach's superior performance and its consistent preference in comparative analyses.

\section{Conclusion}
We leverage multimodal LLMs to introduce MedXChat, a unified intelligent dialogue model supporting the generation of CXR images and reports and intelligent answering based on users' input. MedXChat especially excels in user-prompted generation of fine-grained CXR patterns. It surpasses other SOTA models in multimodal generation across multiple medical imaging tasks. Yet, MedXChat is still \underline{limited} in lacking the capacity for medical visual grounding, another essential medical multimodal task.
\label{sec:conclusion}

\clearpage  % TODO REVIEW/FINAL: This \clearpage needs to be removed from both review and camera-ready versions.

\bibliographystyle{splncs04}
\bibliography{main}

\begin{thebibliography}{10}
\providecommand{\url}[1]{\texttt{#1}}
\providecommand{\urlprefix}{URL }
\providecommand{\doi}[1]{https://doi.org/#1}

\bibitem{ambati2018sequence}
Ambati, R., Dudyala, C.R.: A sequence-to-sequence model approach for imageclef 2018 medical domain visual question answering. In: INDICON (2018)

\bibitem{banerjee2005meteor}
Banerjee, S., Lavie, A.: Meteor: An automatic metric for mt evaluation with improved correlation with human judgments. In: ACL workshop (2005)

\bibitem{baur1804melanogans}
Baur, C., Albarqouni, S., Navab, N.: Melanogans: High resolution skin lesion synthesis with gans. In: arXiv preprint arXiv:1804.04338 (1804)

\bibitem{brown2020language}
Brown, T., Mann, B., Ryder, N., Subbiah, M., et~al.: Language models are few-shot learners. In: NeurIPS (2020)

\bibitem{chambon2022adapting}
Chambon, P., Bluethgen, C., Langlotz, C.P., Chaudhari, A.: Adapting pretrained vision-language foundational models to medical imaging domains. In: NeurIPS (2022)

\bibitem{chen2022exploiting}
Chen, Z., Gu, S., Lu, G., Xu, D.: Exploiting intra-slice and inter-slice redundancy for learning-based lossless volumetric image compression. IEEE Transactions on Image Processing  \textbf{31},  1697--1707 (2022)

\bibitem{chen2022cross}
Chen, Z., Shen, Y., Song, Y., Wan, X.: Cross-modal memory networks for radiology report generation. In: ACL (2022)

\bibitem{chen2020generating}
Chen, Z., Song, Y., Chang, T.H., Wan, X.: Generating radiology reports via memory-driven transformer. In: ACL (2020)

\bibitem{chiang2023vicuna}
Chiang, W.L., Li, Z., Lin, Z., Sheng, Y., Wu, Z., Zhang, H., Zheng, L., Zhuang, S., Zhuang, Y., Gonzalez, J.E., et~al.: Vicuna: An open-source chatbot impressing gpt-4 with 90\%* chatgpt quality (2023)

\bibitem{chuquicusma2018fool}
Chuquicusma, M.J., Hussein, S., Burt, J., Bagci, U.: How to fool radiologists with generative adversarial networks? a visual turing test for lung cancer diagnosis. In: ISBI (2018)

\bibitem{cornia2020meshed}
Cornia, M., Stefanini, M., Baraldi, L., Cucchiara, R.: Meshed-memory transformer for image captioning. In: CVPR (2020)

\bibitem{do2021multiple}
Do, T., Nguyen, B.X., Tjiputra, E., Tran, M., Tran, Q.D., Nguyen, A.: Multiple meta-model quantifying for medical visual question answering. In: MICCAI (2021)

\bibitem{eslami2021does}
Eslami, S., de~Melo, G., Meinel, C.: Does clip benefit visual question answering in the medical domain as much as it does in the general domain? In: arXiv preprint arXiv:2112.13906 (2021)

\bibitem{esser2021taming}
Esser, P., Rombach, R., Ommer, B.: Taming transformers for high-resolution image synthesis. In: CVPR (2021)

\bibitem{finn2017model}
Finn, C., Abbeel, P., Levine, S.: Model-agnostic meta-learning for fast adaptation of deep networks. In: ICML (2017)

\bibitem{frid2018synthetic}
Frid-Adar, M., Klang, E., Amitai, M., Goldberger, J., Greenspan, H.: Synthetic data augmentation using gan for improved liver lesion classification. In: ISBI (2018)

\bibitem{ge2023making}
Ge, Y., Zhao, S., Zeng, Z., Ge, Y., Li, C., Wang, X., Shan, Y.: Making llama see and draw with seed tokenizer. In: arXiv preprint arXiv:2310.01218 (2023)

\bibitem{goodfellow2020generative}
Goodfellow, I., Pouget-Abadie, J., Mirza, M., Xu, B., Warde-Farley, D., Ozair, S., Courville, A., Bengio, Y.: Generative adversarial networks. In: ACM (2020)

\bibitem{he2015deep}
He, K., Zhang, X., Ren, S., Sun, J.: Deep residual learning for image recognition. In: CVPR (2015)

\bibitem{ho2020denoising}
Ho, J., Jain, A., Abbeel, P.: Denoising diffusion probabilistic models. In: NeurIPS (2020)

\bibitem{hou2023diversity}
Hou, Z., Yan, R., Wang, Q., Lang, N., Zhou, X.: Diversity-preserving chest radiographs generation from reports in one stage. In: MICCAI (2023)

\bibitem{hu2021lora}
Hu, E.J., Shen, Y., Wallis, P., Allen-Zhu, Z., Li, Y., Wang, S., Wang, L., Chen, W.: Lora: Low-rank adaptation of large language models. In: ICLR (2021)

\bibitem{jin2023promptmrg}
Jin, H., Che, H., Lin, Y., Chen, H.: Promptmrg: Diagnosis-driven prompts for medical report generation. In: arXiv preprint arXiv:2308.12604 (2023)

\bibitem{jing2017automatic}
Jing, B., Xie, P., Xing, E.: On the automatic generation of medical imaging reports. In: ACL (2017)

\bibitem{johnson2019mimic}
Johnson, A.E., Pollard, T.J., Greenbaum, N.R., Lungren, M.P., Deng, C.y., Peng, Y., Lu, Z., Mark, R.G., Berkowitz, S.J., Horng, S.: Mimic-cxr-jpg, a large publicly available database of labeled chest radiographs. In: arXiv preprint arXiv:1901.07042 (2019)

\bibitem{khare2021mmbert}
Khare, Y., Bagal, V., Mathew, M., Devi, A., Priyakumar, U.D., Jawahar, C.: Mmbert: Multimodal bert pretraining for improved medical vqa. In: ISBI (2021)

\bibitem{kingma2013auto}
Kingma, D.P., Welling, M.: Auto-encoding variational bayes. In: ICLR (2013)

\bibitem{koh2023generating}
Koh, J.Y., Fried, D., Salakhutdinov, R.: Generating images with multimodal language models. In: NeurIPS (2023)

\bibitem{lee2023unified}
Lee, H., Kim, W., Kim, J.H., Kim, T., Kim, J., Sunwoo, L., Choi, E.: Unified chest x-ray and radiology report generation model with multi-view chest x-rays. In: arXiv preprint arXiv:2302.12172 (2023)

\bibitem{lee2023llmcxr}
Lee, S., Kim, W.J., Chang, J., Ye, J.C.: Llm-cxr: Instruction-finetuned llm for cxr image understanding and generation. In: ICLR (2023)

\bibitem{li2019knowledge}
Li, C.Y., Liang, X., Hu, Z., Xing, E.P.: Knowledge-driven encode, retrieve, paraphrase for medical image report generation. In: AAAI (2019)

\bibitem{li2024llava}
Li, C., Wong, C., Zhang, S., Usuyama, N., Liu, H., Yang, J., Naumann, T., Poon, H., Gao, J.: Llava-med: Training a large language-and-vision assistant for biomedicine in one day. NIPS  (2024)

\bibitem{lin2004rouge}
Lin, C.Y.: Rouge: A package for automatic evaluation of summaries. In: Text summarization branches out (2004)

\bibitem{liu2021exploring}
Liu, F., Wu, X., Ge, S., Fan, W., Zou, Y.: Exploring and distilling posterior and prior knowledge for radiology report generation. In: CVPR (2021)

\bibitem{lu2017knowing}
Lu, J., Xiong, C., Parikh, D., Socher, R.: Knowing when to look: Adaptive attention via a visual sentinel for image captioning. In: CVPR (2017)

\bibitem{nguyen2019overcoming}
Nguyen, B.D., Do, T.T., Nguyen, B.X., Do, T., Tjiputra, E., Tran, Q.D.: Overcoming data limitation in medical visual question answering. In: MICCAI (2019)

\bibitem{openai2023gpt4}
OpenAI: Gpt-4 technical report (2023)

\bibitem{papineni2002bleu}
Papineni, K., Roukos, S., Ward, T., Zhu, W.J.: Bleu: a method for automatic evaluation of machine translation. In: ACL (2002)

\bibitem{peng2023instruction}
Peng, B., Li, C., He, P., Galley, M., Gao, J.: Instruction tuning with gpt-4. In: arXiv preprint arXiv:2304.03277 (2023)

\bibitem{radford2021learning}
Radford, A., Kim, J.W., Hallacy, C., Ramesh, A., et~al.: Learning transferable visual models from natural language supervision. In: ICML (2021)

\bibitem{radford2018improving}
Radford, A., Narasimhan, K., Salimans, T., Sutskever, I., et~al.: Improving language understanding by generative pre-training. OpenAI (2018)

\bibitem{rennie2017self}
Rennie, S.J., Marcheret, E., Mroueh, Y., Ross, J., Goel, V.: Self-critical sequence training for image captioning. In: CVPR (2017)

\bibitem{rombach2022high}
Rombach, R., Blattmann, A., Lorenz, D., Esser, P., Ommer, B.: High-resolution image synthesis with latent diffusion models. In: CVPR (2022)

\bibitem{ronneberger2015u}
Ronneberger, O., Fischer, P., Brox, T.: U-net: Convolutional networks for biomedical image segmentation. In: MICCAI (2015)

\bibitem{tanida2023interactive}
Tanida, T., M{\"u}ller, P., Kaissis, G., Rueckert, D.: Interactive and explainable region-guided radiology report generation. In: CVPR (2023)

\bibitem{taori2023stanford}
Taori, R., Gulrajani, I., Zhang, T., Dubois, Y., Li, X., Guestrin, C., Liang, P., Hashimoto, T.B.: Stanford alpaca: an instruction-following llama model (2023) (2023)

\bibitem{thawkar2023xraygpt}
Thawkar, O., Shaker, A., Mullappilly, S.S., Cholakkal, H., Anwer, R.M., Khan, S., Laaksonen, J., Khan, F.S.: Xraygpt: Chest radiographs summarization using medical vision-language models. In: arXiv preprint arXiv:2306.07971 (2023)

\bibitem{touvron2023llama}
Touvron, H., Lavril, T., Izacard, G., Martinet, X., Lachaux, M.A., Lacroix.: Llama: Open and efficient foundation language models. In: arXiv preprint arXiv:2302.13971 (2023)

\bibitem{vaswani2017attention}
Vaswani, A., Shazeer, N., Parmar, N., Uszkoreit, J., Jones, L., Gomez, A.N., Kaiser, {\L}., Polosukhin, I.: Attention is all you need. In: NeurIPS (2017)

\bibitem{vedantam2015cider}
Vedantam, R., Lawrence~Zitnick, C., Parikh, D.: Cider: Consensus-based image description evaluation. In: CVPR (2015)

\bibitem{vinyals2015show}
Vinyals, O., Toshev, A., Bengio, S., Erhan, D.: Show and tell: A neural image caption generator. In: CVPR (2015)

\bibitem{wang2022automated}
Wang, Z., Han, H., Wang, L., Li, X., Zhou, L.: Automated radiographic report generation purely on transformer: A multicriteria supervised approach. In: IEEE Transactions on Medical Imaging (2022)

\bibitem{wang2022medical}
Wang, Z., Tang, M., Wang, L., Li, X., Zhou, L.: A medical semantic-assisted transformer for radiographic report generation. In: MICCAI (2022)

\bibitem{wang2021self}
Wang, Z., Zhou, L., Wang, L., Li, X.: A self-boosting framework for automated radiographic report generation. In: CVPR (2021)

\bibitem{wang2022medclip}
Wang, Z., Wu, Z., Agarwal, D., Sun, J.: Medclip: Contrastive learning from unpaired medical images and text. In: ACL (2022)

\bibitem{wolleb2022diffusion}
Wolleb, J., Bieder, F., Sandk{\"u}hler, R., Cattin, P.C.: Diffusion models for medical anomaly detection. In: MICCAI (2022)

\bibitem{wu2023towards}
Wu, C., Zhang, X., Zhang, Y., Wang, Y., Xie, W.: Towards generalist foundation model for radiology. arXiv preprint arXiv:2308.02463  (2023)

\bibitem{wu2023nextgpt}
Wu, S., Fei, H., Qu, L., Ji, W., Chua, T.S.: Next-gpt: Any-to-any multimodal llm. In: arXiv preprint arXiv:2309.05519 (2023)

\bibitem{xu2023elixr}
Xu, S., Yang, L., Kelly, C., Sieniek, M., et~al.: Elixr: Towards a general purpose x-ray artificial intelligence system through alignment of large language models and radiology vision encoders. In: arXiv preprint arXiv:2308.01317 (2023)

\bibitem{yang2022knowledge}
Yang, S., Wu, X., Ge, S., Zhou, S.K., Xiao, L.: Knowledge matters: Chest radiology report generation with general and specific knowledge. In: Medical image analysis (2022)

\bibitem{ye2023mplug}
Ye, Q., Xu, H., Xu, G., Ye, J., Yan, M., Zhou, Y., Wang, J., Hu, A., Shi, P., Shi, Y., et~al.: mplug-owl: Modularization empowers large language models with multimodality. arXiv preprint arXiv:2304.14178  (2023)

\bibitem{yin2019automatic}
Yin, C., Qian, B., Wei, J., Li, X., Zhang, X., Li, Y., Zheng, Q.: Automatic generation of medical imaging diagnostic report with hierarchical recurrent neural network. In: ICDM (2019)

\bibitem{you2021aligntransformer}
You, D., Liu, F., Ge, S., Xie, X., Zhang, J., Wu, X.: Aligntransformer: Hierarchical alignment of visual regions and disease tags for medical report generation. In: MICCAI (2021)

\bibitem{zhang2023adding}
Zhang, L., Rao, A., Agrawala, M.: Adding conditional control to text-to-image diffusion models. In: ICCV (2023)

\bibitem{zhang2020radiology}
Zhang, Y., Wang, X., Xu, Z., Yu, Q., Yuille, A., Xu, D.: When radiology report generation meets knowledge graph. In: AAAI (2020)

\end{thebibliography}
\end{document}